\documentclass[10pt,twocolumn,letterpaper]{article}

\usepackage{cvpr}

\usepackage[hyphens]{url}
\usepackage{graphicx}
\usepackage{amsmath}
\usepackage{amssymb}
\usepackage{booktabs}
\usepackage{multirow}

\definecolor{projpink}{RGB}{214,0,143}
\newcommand{\projlink}[2]{\href{#1}{\textcolor{projpink}{\ttfamily\small #2}}}
\definecolor{citecolor}{HTML}{c0392b} % 2980b9
\definecolor{linkcolor}{HTML}{c0392b}
\usepackage[pagebackref,breaklinks,colorlinks,citecolor=citecolor,linkcolor=linkcolor]{hyperref}

\def\paperID{*****}
\def\confName{CVPR}
\def\confYear{2026}

\title{RealityBridge: Bridging Editable 3D Gaussian Splatting Driving\\Simulations and Real-World Videos}
\author{
Zhenhua Wu$^{1,2,*}$ \quad
Yun Pang$^{1,*}$ \quad
Mingkun Chang$^{1,*}$\\
Yuwei Ning$^{1}$ \quad
Liangzhi Wang$^{1}$ \quad
Yi Xiao$^{1}$ \quad
Guanbin Li$^{1,3,\dagger}$\\[0.35em]
Sun Yat-sen University$^{1}$ \quad
Shanghai Innovation Institute$^{2}$\\
Shenzhen Loop Area Institute$^{3}$\\[0.25em]
Project Page:
\projlink{https://arthurwuzh.github.io/RealityBridge/}{arthurwuzh.github.io/RealityBridge}
}

\makeatletter
\apptocmd{\@maketitle}{%
    \begingroup
    \centering
    \includegraphics[width=0.98\textwidth]{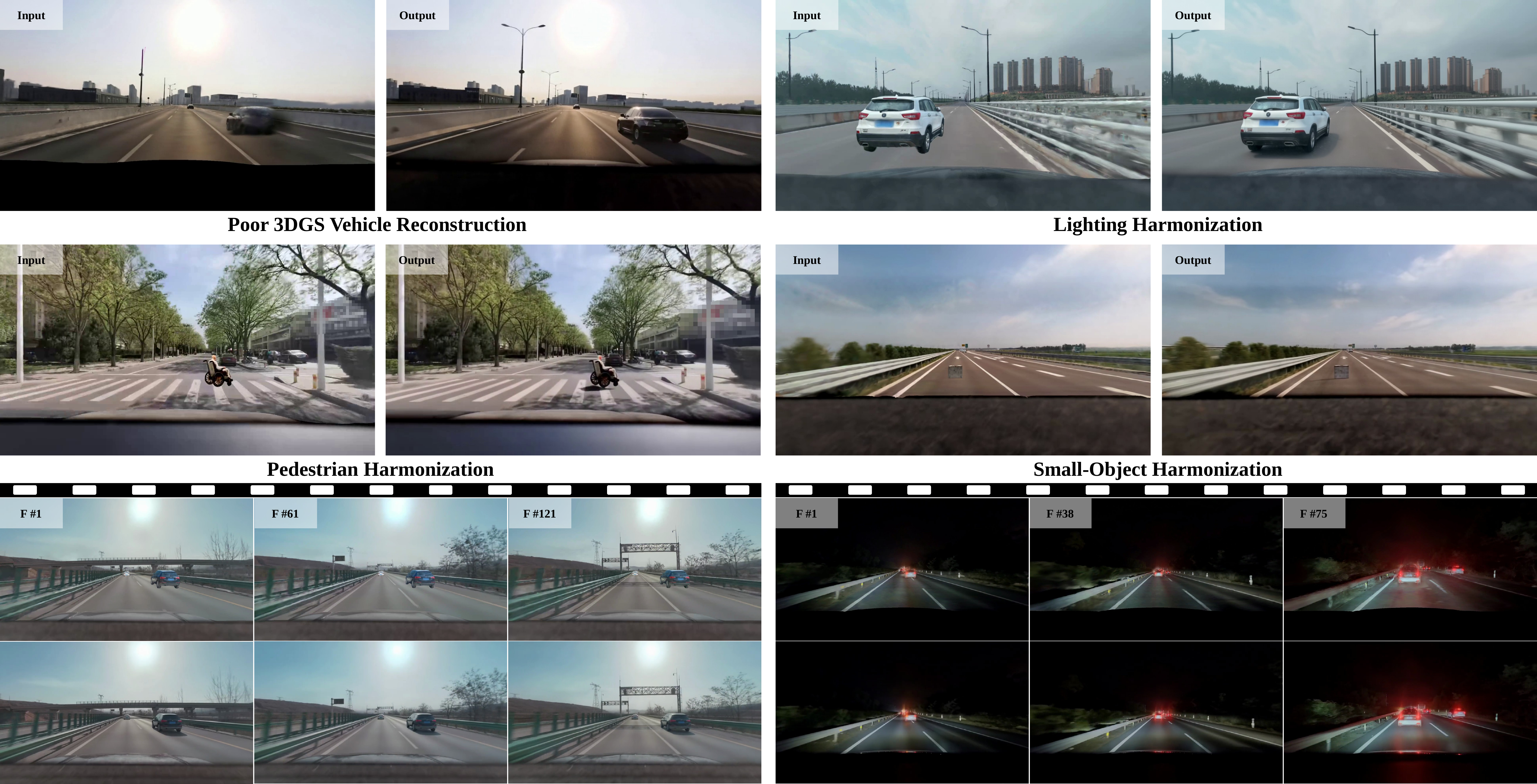}
    \captionsetup{belowskip=15pt}
    \vspace{-0.5em}
    \captionof{figure}{\textbf{Restoration and harmonization results.} RealityBridge improves realism and foreground harmonization while preserving temporal consistency in degraded or edited 3DGS simulation scenarios. \textbf{Top:} four representative cases, including restoration of poor vehicle reconstruction and lighting harmonization for vehicles, pedestrians, and small objects. \textbf{Bottom:} frames sampled at different timestamps from daytime and nighttime scenes, demonstrating stable and consistent results over time.}
    \label{fig:teaser}
    \par
    \endgroup
}{}{}
\makeatother

\begin{document}
\maketitle
\begingroup
\renewcommand{\thefootnote}{}
\footnotetext{$^*$Equal Contribution\hspace{0.2cm} $^\dagger$Corresponding Author}
\endgroup

\begin{abstract}
Long-tail hazardous scenarios are essential for safety-oriented autonomous driving, yet they are difficult to collect at scale. 
Editable 3D Gaussian Splatting (3DGS) simulation offers a scalable alternative through real-scene reconstruction and controllable editing.
However, edited 3DGS-rendered videos often exhibit a significant Sim-to-Real gap, manifested as rendering artifacts, degraded foreground assets, illumination mismatch, and temporal flickering. Addressing these coupled defects requires jointly restoring local appearance, harmonizing edited content, and maintaining temporal consistency, whereas existing methods typically address only a subset of these requirements.
To fill this gap, we propose \textit{\textbf{RealityBridge}}, a video restoration and harmonization framework that converts edited 3DGS renderings into realistic driving footage while preserving simulator-defined structure, edits, and dynamics.
RealityBridge conditions a video foundation model on complementary modality signals, with a lightweight GateNet adaptively controlling their injection across backbone blocks.
We further develop a task-oriented curation pipeline to construct training data, and design a four-stage supervised training strategy followed by reward-guided post-training.
Extensive experiments demonstrate that RealityBridge outperforms existing methods in restoration and harmonization while preserving strong temporal consistency.
\end{abstract}

\section{Introduction}

Safety-oriented training and evaluation of autonomous driving systems require diverse, controllable, and reproducible long-tail hazardous scenarios, which are rare, costly to collect, and difficult to reproduce in the real world. Recent advances in 3D Gaussian Splatting (3DGS)~\cite{3dgs} provide a promising foundation for editable driving simulation~\cite{unisim,mars,suds,neurad,drivinggaussian,streetgaussians,emernerf}. By reconstructing real-world driving scenes and supporting edits such as object insertion, removal, and trajectory modification, 3DGS makes it possible to build editable simulators in which safety-critical objects, trajectories, and interactions can be controlled for training, evaluation, and safety validation.

However, being editable does not necessarily make the simulator realistic. Edited 3DGS-rendered videos still suffer from a Sim-to-Real gap, such as rendering artifacts, degraded foreground assets, inconsistent illumination, and temporal flickering~\cite{ad-simulation-survey, ad-simulation-survey-2}.
These defects introduce domain bias and limit their use in autonomous driving tasks. Therefore, a bridge from edited 3DGS simulations to photorealistic footage must improve visual realism without altering the simulator-defined scene layout, edited assets, motion trajectories, and safety-critical interactions that are essential for scenario generation.

% Existing methods cannot fully convert editable simulation into photorealistic videos. 
Existing methods address only partial aspects of translating editable simulations into photorealistic videos.
Image-prior-based approaches~\cite{ganerf,reconfusion,nerfbusters,difix3d,gsfix3d,freefix,d3dr,r3d2} can alleviate local 3DGS rendering artifacts, but they are often applied frame by frame and lack temporal modeling for continuous driving videos. They are also not designed to jointly handle background restoration, foreground harmonization, and structure preservation. Video-prior-based methods~\cite{cat3d,viewcrafter,3dgs-enhancer,gsfixer,genfusion,gaussfusion,diff-harmonizer,cosmos} provide stronger temporal and generative priors, but mainly target generic scene repair, video editing, or cross-domain style transfer rather than editable 3DGS driving simulation. 
As a result, the problem remains open: how to jointly restore and harmonize edited 3DGS videos while preserving simulator-defined structure, edits, and dynamics.
% existing methods fail to simultaneously improve visual fidelity, preserve simulator-defined edits, and maintain temporal consistency.

To fill the gap, we propose \textit{\textbf{RealityBridge}}, a post-rendering refinement framework that converts edited 3DGS-rendered simulations into photorealistic videos while preserving simulator-defined structure, edits, and dynamics. At its core, RealityBridge adopts a video-based refinement model that performs restoration and harmonization built on a video generation backbone. Multimodal control signals provide spatial, structural, and semantic guidance, enabling the model to identify regions that require restoration or harmonization, safety-critical structures that should be preserved, and object-specific realism requirements. Since these heterogeneous controls vary in importance across generation stages and backbone layers, we introduce a lightweight GateNet to adaptively modulate their layer-wise control features, balancing structural fidelity and realistic appearance synthesis. Finally, targeted data curation, a four-stage supervised training curriculum supporting chunk-wise autoregressive processing, and reward-guided post-training are incorporated to improve realism, temporal stability, and simulator fidelity.

The main contributions are summarized as follows:

\begin{itemize}

\item[$\bullet$]\textbf{Video-based Refinement Model.}
We propose RealityBridge, a video-based refinement model that adaptively modulates layer-wise ControlNet features for translating edited 3DGS simulations into photorealistic footage.

\item[$\bullet$]\textbf{Targeted Data Curation Pipeline.}
We develop a task-oriented pipeline to curate Sim-to-Real data covering artifacts, relighting, human motion, and small objects, with multimodal signals for fine-grained supervision.

\item[$\bullet$]\textbf{Extensive Experimental Validation.}
Our qualitative and quantitative results show that RealityBridge outperforms existing methods in artifact restoration, illumination harmonization, and temporal consistency.

\end{itemize}

\section{Related Work}
\begin{figure*}[t]
    \centering
    \includegraphics[width=1.0\textwidth]{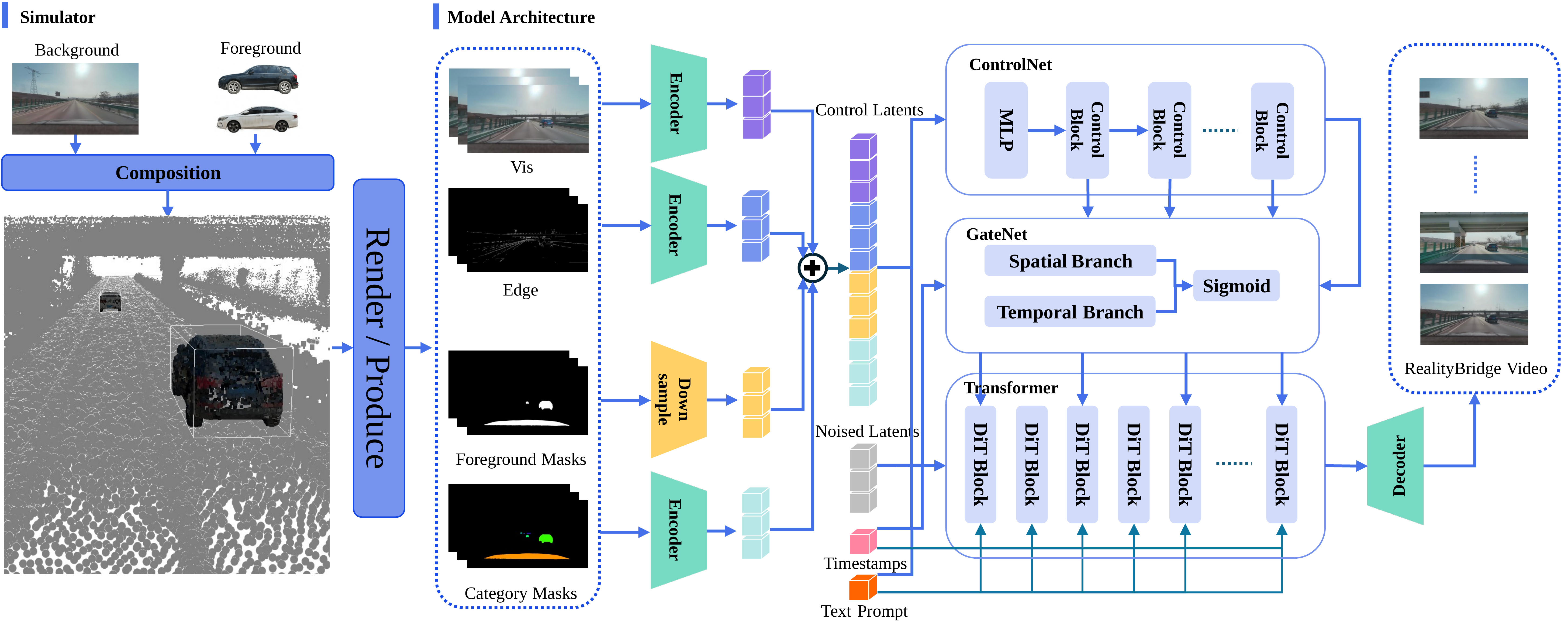}
    \vspace{-1.2em}
    \caption{
\textbf{RealityBridge model pipeline at a glance.}
The RealityBridge model converts edited 3DGS simulation videos into photorealistic footage. Multimodal conditions, including visual inputs, edge maps, and category masks, are processed by ControlNet into layer-wise control features. GateNet then adaptively modulates these features before they are injected into the flow-transformer backbone, enabling region-aware restoration and harmonization. 
}
    \label{fig:pipeline1}
    \vspace{-1.2em}
\end{figure*}

\noindent\textbf{Editable Neural Driving Simulators.}\quad
Recent 3DGS-based neural simulators~\cite{3dgs,drivinggaussian,streetgaussians} support controllable driving-scene editing, such as object insertion, trajectory modification, and extrapolated-view rendering~\cite{hugsim,splatad,psd-simulation,xsim,lidar-evs,unisim,mars,suds,neurad,emernerf}. However, such interventions often disrupt the geometric and photometric consistency of the reconstructed scenes~\cite{sugur,gs-editor}, leading to rendering artifacts, inconsistent illumination, and temporal flickering. % These degradation patterns reveal a persistent Sim-to-Real gap, motivating post-rendering refinement methods that improve realism while preserving the intended edits.
As a result, post-rendering refinement remains a key barrier before editable 3DGS-based neural simulators can become a production-ready option for large-scale driving simulation.

\noindent\textbf{Post-rendering Refinement with Image Priors.}\quad
Recent works leverage image generative priors to mitigate rendering artifacts~\cite{ganerf,reconfusion,nerfbusters}. Difix3D+~\cite{difix3d} uses a single-step diffusion model to eliminate novel-view defects in 3DGS renderings. Extending this, GSFix3D~\cite{gsfix3d} fine-tunes a latent diffusion model on scene-specific data to repair missing regions, while FreeFix~\cite{freefix} corrects extrapolated views and updates underlying representations. Based on Difix3D+, NVIDIA Fixer~\footnote{\url{https://github.com/nv-tlabs/Fixer}} introduces a neural enhancer tailored for simulation observations to improve visual fidelity. 
For asset harmonization, D3DR~\cite{d3dr} and R3D2~\cite{r3d2} use diffusion guidance to generate realistic illumination and shadows for inserted 3D assets. While effective for localized static enhancement, their frame-wise 2D formulations lack the spatiotemporal modeling required to jointly preserve temporal coherence, object-scene consistency, and simulator-defined structure in edited 3DGS driving videos.

\noindent\textbf{Post-rendering Refinement with Video Priors.}\quad
For general 3DGS enhancement, 3DGS-Enhancer~\cite{3dgs-enhancer} and GSFixer~\cite{gsfixer} adopt video generation backbones to restore artifact-prone novel views~\cite{cat3d,viewcrafter}. Incorporating spatial cues, GenFusion~\cite{genfusion} and GaussFusion~\cite{gaussfusion} condition on RGB-D inputs or Gaussian buffers to suppress temporal flickering and floating artifacts.
For video harmonization, Lumen~\cite{zeng2025lumen} relights human foreground subjects and harmonizes them with new backgrounds. DiffusionHarmonizer~\cite{diff-harmonizer} improves appearance harmonization for neural-rendered driving scenes, and Cosmos-Transfer2.5~\cite{cosmos} provides general visual style transfer under flexible conditioning. 
Although their video priors improve spatial and temporal consistency, these methods cannot jointly achieve restoration and harmonization while preserving simulator-defined structure, edits, and dynamics.

\section{Methodology}

\textit{\textbf{RealityBridge}} comprises four key components: a video-based refinement model with adaptive multimodal control, a targeted data curation pipeline, a progressive chunk-wise training strategy, and reward-guided post-training.

\subsection{Model Architecture of RealityBridge}

\subsubsection{Problem Formulation.}
We formulate post-rendering refinement of edited 3DGS simulations as a multimodal video-to-video restoration and harmonization task. Given a 3DGS-rendered simulation video $V^{sim}$ and multimodal conditions $C$, the model generates a photorealistic video $\hat{V}^{real}$:
\begin{equation}
    \hat{V}^{real} = \mathcal{F}_{\theta}(V^{sim}, C).
\end{equation}
We instantiate $\mathcal{F}_{\theta}$ as a conditional latent video generation model. Different from generic video generation, the model conditions on the edited 3DGS rendering and multimodal controls to enhance visual realism while preserving simulator-defined scene content. Accordingly, the task involves three coupled objectives: removing rendering artifacts, harmonizing foreground assets with the background, and maintaining temporal consistency across driving videos. To address these, RealityBridge adopts the multimodal gated architecture illustrated in Fig.~\ref{fig:pipeline1}.

\subsubsection{Multimodal Condition Encoding.}
The edited 3DGS video alone does not disambiguate which regions require restoration, harmonization, or preservation. Different scene regions demand distinct corrections, while key geometric and semantic structures should remain unchanged. Accordingly, we introduce multimodal control signals to provide explicit regional, structural, and semantic guidance for restoration and harmonization.

\begin{figure*}[t]
    \centering
    \includegraphics[width=0.96\textwidth]{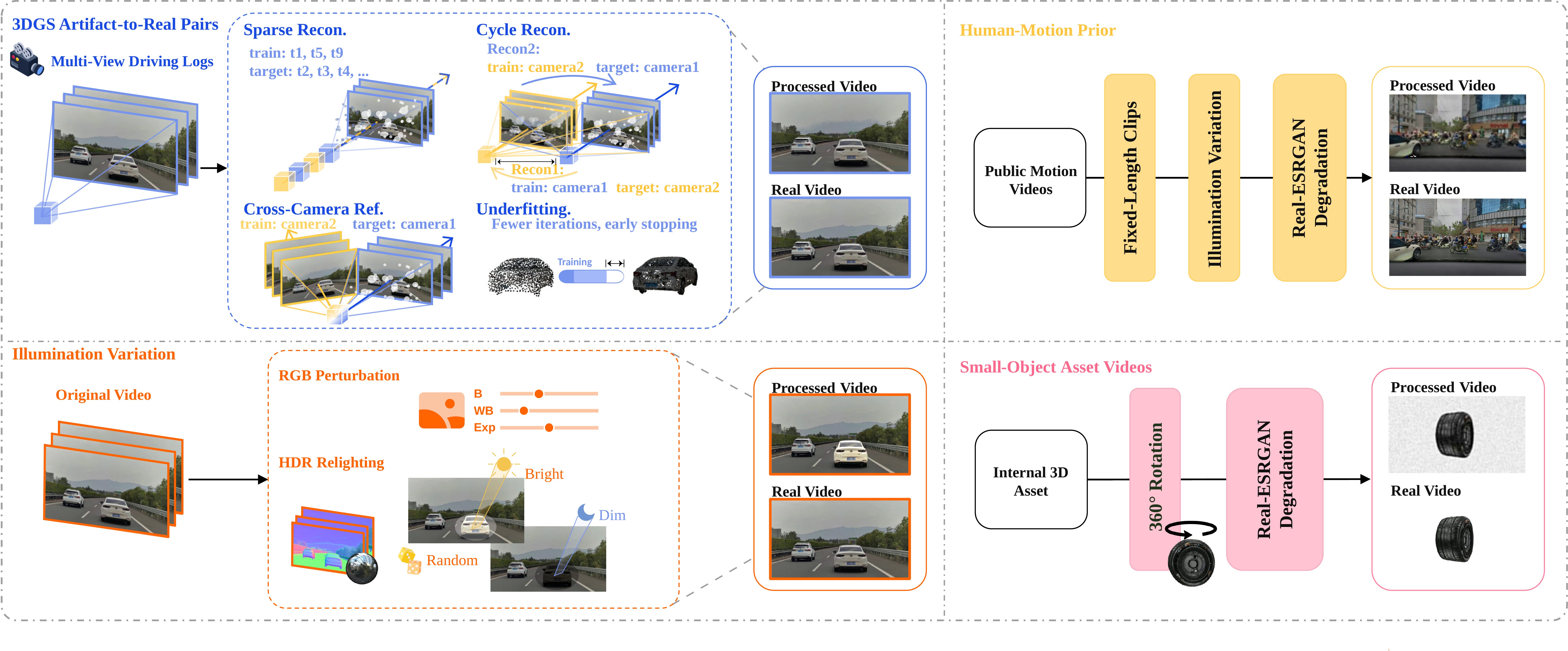}
    \vspace{-1.25em}
    \caption{\textbf{Overview of the data pipeline.} We construct paired supervision from 3DGS Artifact-to-Real pairs, illumination variation, human-motion priors, and small-object asset videos.}
    \vspace{-1.2em}
    \label{fig:data_curation_pipeline}
\end{figure*}

Specifically, the 3DGS-rendered video latent $z_{sim}$ provides the global layout, camera motion, and appearance reference. 
The foreground-mask $M_f$ is constructed by taking the element-wise union of frame-wise category masks for selected object classes, including vehicles, pedestrians, and small objects. It is then patchified and resized to match the video latents, yielding the foreground-mask condition $\widetilde{M}_f$, which provides class-agnostic localization of task-relevant foreground regions. The VAE-encoded edge representation $z_E$ guides the preservation of structural boundaries and safety-critical details.
In contrast, the category masks are represented as a semantic label map and encoded through the video VAE to obtain $z_M$, providing category-aware semantic priors for the generation process. These modalities are concatenated in a fixed order, yielding the unified condition tensor $z_C$:
\begin{equation}
    z_C = \text{Concat}(z_{sim}, z_E, \widetilde{M}_f, z_M),
\end{equation}
The ControlNet branch~\cite{controlnet} projects the unified condition tensor into control features:
\begin{equation}
    F_c = \Phi_{\text{ControlNet}}(z_C).
\end{equation}
The resulting control features guide the DiT backbone to perform restoration and harmonization in the latent space.

\subsubsection{GateNet for Adaptive Control Injection.}
After concatenation, the multimodal conditions are processed by ControlNet to produce layer-wise control features $\{F_c^{(l)}\}_{l=1}^{L}$. Standard additive injection directly adds each $F_c^{(l)}$ to the corresponding DiT block without adapting its contribution to the current generation state. We instead introduce GateNet to perform layer-specific adaptive control injection. At each injection layer, GateNet predicts an adaptive injection weight based on the DiT hidden state, the corresponding control feature, and the diffusion timestep.

For injection layer $l$, let
$h^{(l)},F_c^{(l)}\in\mathbb{R}^{B\times S\times D}$
denote the aligned DiT and ControlNet token sequences, respectively, where $B$ is the batch size, $S$ is the number of flattened spatiotemporal latent tokens, and $D$ is the token feature dimension. The spatial branch $\mathcal{G}_{\eta,l}^{s}$ computes one injection logit for each token from their concatenated features:
\begin{equation}
s^{(l)}
=
\mathcal{G}_{\eta,l}^{s}
\left(
\operatorname{Concat}
\left[
h^{(l)}, F_c^{(l)}
\right]
\right),
\end{equation}
where $s^{(l)}\in\mathbb{R}^{B\times S\times 1}$ denotes an injection logit for each latent token.
To adapt the injection strength during denoising, a layer-specific temporal branch $\mathcal{G}_{\eta,l}^{t}$ projects the timestep modulation vector $e_t\in\mathbb{R}^{B\times D}$ into an additive bias:
\begin{equation}
b_t^{(l)}
=
\mathcal{G}_{\eta,l}^{t}
\left(
e_t
\right),
\end{equation}
Here 
$b_t^{(l)}\in\mathbb{R}^{B\times 1\times 1}$ is the layer-specific timestep bias. The adaptive injection gate is then computed as
\begin{equation}
\gamma^{(l)}
=
\sigma
\left(
s^{(l)} + b_t^{(l)}
\right),
\end{equation}
where $\sigma(\cdot)$ is the sigmoid function and $\gamma^{(l)}$ denotes the adaptive injection gate at the $l$-th DiT layer.
The gated control feature is injected into the corresponding DiT block as
\begin{equation}
\widetilde{h}^{(l)}
=
h^{(l)}
+
\gamma^{(l)}
\odot
F_c^{(l)},
\end{equation}
where $\odot$ denotes token-wise scaling over the feature dimension and $\widetilde{h}^{(l)}$ denotes the updated state at injection point $l$.

\subsection{Targeted Data Curation}
% Training RealityBridge model requires supervision aligned with its restoration and harmonization objectives. Since edited 3DGS videos contain diverse failures, we develop a data curation pipeline for targeted 3DGS-to-Real supervision, as shown in Fig.~\ref{fig:data_curation_pipeline}. 

Training the RealityBridge model requires supervision aligned with its restoration and harmonization objectives. Since edited 3DGS videos contain diverse failures, we develop a data curation pipeline for targeted 3DGS-to-Real supervision, illustrated in Fig.~\ref{fig:data_curation_pipeline} and detailed in Appendix~\ref{app:data_curation}.
    
    \textbf{3DGS Artifact-to-Real Pairs.} We construct 3DGS Artifact-to-Real pairs from internal multi-view driving videos covering highways, rural roads, urban streets, and garages. Following Difix3D+~\cite{difix3d}, we generate degraded 3DGS renderings through sparse reconstruction, cycle reconstruction, cross-camera reference, and underfitting, providing supervision for Artifact-to-Real mapping.

    \textbf{Illumination Variation.} We construct illumination-variation data to supervise foreground harmonization. Foreground regions are perturbed at the RGB level and relighted using normal-guided lighting inspired by Lumen~\cite{zeng2025lumen}. The original videos are used as supervision, enabling illumination correction, shadow recovery, and boundary blending.

    \textbf{Human-Motion Prior.} 
    We select videos captured from stationary vehicles featuring multiple pedestrians as human-motion priors, covering diverse motions, poses, occlusions, and interactions. Degraded counterparts are synthesized using the degradation pipeline from Real-ESRGAN~\cite{realesrgan} along with illumination perturbation to enhance robustness in human-region restoration.

    \textbf{Small-Object Asset Videos.} We curate object-centric videos for long-tail safety-critical obstacles, such as traffic signs, cones, barriers, and fallen objects. Since these objects are rare and easily degraded in 3DGS reconstruction, we use 360-degree captures from our internal 3D asset pipeline for small-object restoration and preservation.

\subsection{Four-stage Supervised Training}
\label{sec:training_strategy}

\textbf{Training Pipeline.}
RealityBridge generates long videos autoregressively by processing them as a sequence of chunks. The first chunk is generated without historical context, whereas each subsequent chunk is conditioned on preceding latent frames. To support both generation regimes while progressively introducing multimodal control, we adopt a four-stage training pipeline. 
During training, we define the \emph{context frames} as the leading latent frames encoded from the target real video. When provided, they serve as historical context for temporal continuation.

\begin{enumerate}
    \item[\textit{i.}] \textit{Context warm-up.}
   We train DiT with one or two context frames to learn continuation from historical context. 

    \item[\textit{ii.}] \textit{Context-frame dropout.}
    We randomly provide zero, one, or two context frames, enabling both context-free initialization and context-conditioned continuation.

    \item[\textit{iii.}] \textit{Adaptive control learning.}
    We freeze DiT and train ControlNet and GateNet to incorporate multimodal conditions without disrupting the learned temporal prior.

    \item[\textit{iv.}] \textit{Joint refinement.}
    We jointly fine-tune the DiT backbone, ControlNet, and GateNet to coordinate autoregressive temporal generation with adaptive multimodal control.
\end{enumerate}

\paragraph{Training Objective.}
We adopt conditional flow matching~\cite{flow1-straight-fast-learning,flow2-building-normalizing-flows-stochastic} to optimize the video generation backbone. Given a training tuple $(V^{sim},V^{real},C)$, the video VAE encoder $\mathcal{E}$ maps the target real video to the data latent $z_0=\mathcal{E}(V^{real})$. We sample a Gaussian noise latent $z_1\sim\mathcal{N}(0,I)$ and a timestep $t\sim\mathcal{U}(0,1)$, and construct the intermediate latent $z_t=(1-t)z_1+t z_0$, with a target velocity of $v^*=z_0-z_1$. Conditioned on the intermediate latent, timestep, 3DGS-rendered video, and multimodal conditions, the model predicts the following:
\begin{equation}
    \hat{v}_{\theta}
    =
    v_{\theta}(z_t,t,V^{sim},C).
\end{equation}
The standard flow-matching loss is defined as:
\begin{equation}
    \mathcal{L}_{fm}
    =
    \mathbb{E}\!\left[
    \left\|\hat{v}_{\theta}-v^*\right\|_2^2
    \right].
\end{equation}
Since task-relevant foreground regions occupy only a small portion of the video yet typically require stronger restoration and harmonization, the standard flow-matching loss does not distinguish them from background regions. We therefore introduce a foreground-region loss:
\begin{equation}
    \mathcal{L}_{reg}
    =
    \mathbb{E}\!\left[
    \left\|\widetilde{M}_f\odot(\hat{v}_{\theta}-v^*)\right\|_2^2
    \right].
\end{equation}
We optimize the final objective:
\begin{equation}
    \mathcal{L}
    =
    \mathcal{L}_{fm}
    +
    \lambda_{reg}\mathcal{L}_{reg},
\end{equation}
where the foreground mask $\widetilde{M}_f$ emphasizes restoration and harmonization of vehicles, pedestrians, and small objects, while $\lambda_{reg}$ controls the regional reweighting strength.

\subsection{Reward-guided Post-training}
Supervised training enables effective restoration and harmonization, while distant regions in challenging scenes may occasionally exhibit minor structural deviations during realism enhancement. We therefore introduce reward-guided post-training~\cite{flow-dpo} as a final alignment step to further align photorealism with simulator constraints.

Given an input rendered sequence $V^{sim}=\{I^{r}_{t}\}_{t=1}^{N}$ and a generated video $\hat{V}=\{\hat{I}_{t}\}_{t=1}^{N}$, rewards are computed on a sparsely sampled frame set $\mathcal{S}$. The reward combines an aesthetic term for visual realism and a bounding-box IoU term for foreground structure preservation:
\begin{equation}
\begin{aligned}
R ={}&
\lambda_{\mathrm{aes}}
\mathcal{N}\!\left(
f_{\mathrm{aes}}\!\left(
\{\hat{I}_t\}_{t\in\mathcal{S}}
\right)\right) \\
&+
\lambda_{\mathrm{box}}
\mathcal{N}\!\left(
\sum_{t\in\mathcal{S}}
\frac{1}{|\mathcal{S}|}
\operatorname{IoU}\!\left(
\mathcal{B}(\hat{I}_t),
\mathcal{B}(I_t^r)
\right)\right).
\end{aligned}
\end{equation}
Here, $\lambda_{\mathrm{aes}}$ and $\lambda_{\mathrm{box}}$ weight the aesthetic and bounding-box rewards, respectively. $f_{\mathrm{aes}}$ denotes the aesthetic reward model, $\mathcal{B}(\cdot)$ is an off-the-shelf object detector~\cite{yolov11}, and $\mathcal{N}(\cdot)$ denotes z-score normalization. By computing rewards on sparsely sampled frames, this stage covers different temporal positions in videos while controlling training cost.
More details are provided in Appendix~\ref{app:training_strategy}.

\begin{figure*}[t!]
    \centering
    \includegraphics[width=0.98\textwidth]{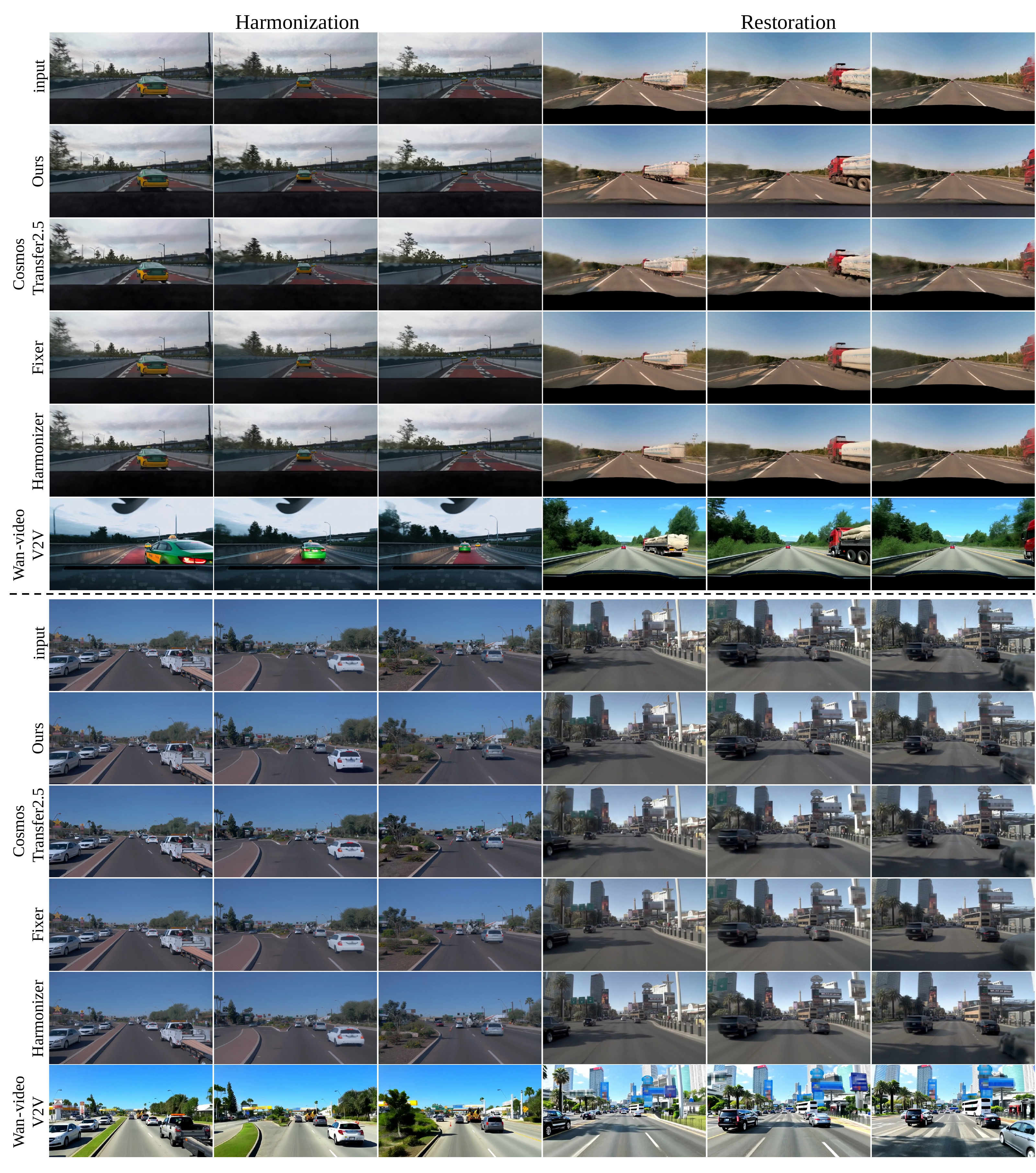}
    \vspace{-1.25em}
    \caption{
        \textbf{Comparison of harmonization and restoration tasks.} We compare different methods on public and internal datasets. 
        % Our method produces more natural illumination and shadow effects in the harmonization task, removes 3DGS artifacts more effectively in the restoration task, and maintains better temporal consistency than the compared methods.
    }
    \label{fig:qualitative_eval}
    \vspace{-1.2em}
\end{figure*}

\section{Experiments}
\begin{table*}[t!]
\centering

\resizebox{\textwidth}{!}{%
\begin{tabular}{c|cccccc|cccccccc}
\toprule
\multirow{2}{*}[-0.7ex]{\textbf{Method}} &
\multicolumn{6}{c|}{\textbf{Harmonization}} &
\multicolumn{8}{c}{\textbf{Restoration}} \\
\cmidrule(lr){2-7} \cmidrule(lr){8-15}
&
\textbf{FID} $\downarrow$ &
\textbf{FVD} $\downarrow$ &
\textbf{SC} $\uparrow$ &
\textbf{TF} $\uparrow$ &
\textbf{MS} $\uparrow$ &
\textbf{DS} $\downarrow$ &
\textbf{FID} $\downarrow$ &
\textbf{FVD} $\downarrow$ &
\textbf{SC} $\uparrow$ &
\textbf{TF} $\uparrow$ &
\textbf{MS} $\uparrow$ &
\textbf{DS} $\downarrow$ &
\textbf{PSNR} $\uparrow$ &
\textbf{SSIM} $\uparrow$ \\
\midrule

SDEdit (SD 3) &
84.17 &
2213.22 &
0.9088 &
0.9654 &
0.9714 &
0.5003 &
\underline{44.53} &
753.65 &
0.8749 &
0.9709 &
0.9769 &
0.7446&
21.35 &
0.679 \\

InstructPix2Pix &
85.96 &
2149.62 &
0.8685 &
0.9351 &
0.9442 &
0.6996 &
69.96 &
1471.39 &
0.8286 &
0.9330 &
0.9434 &
0.9173&
14.30 &
0.488 \\

Fixer &
87.53 &
1426.08 &
0.9458 &
\underline{0.9828} &
\underline{0.9867} &
\underline{0.3091} &
51.86 &
736.85 &
0.9260 &
0.9808 &
0.9861 &
0.6058&
\underline{21.89} &
\underline{0.777} \\

Cosmos-Transfer2.5 &
\textbf{73.21} &
\underline{1344.21} &
0.9424 &
0.9794 &
0.9840 &
0.3341 &
53.91 &
\underline{727.41} &
0.9203 &
0.9792 &
\underline{0.9864} &
0.5538&
21.57 &
0.749 \\

Harmonizer&
84.31&
1364.90&
0.9434&
0.9815&
0.9852&
\textbf{0.2989} &
56.40&
828.25&
0.9111&
\underline{0.9809}&
0.9850&
\underline{0.5303}&
21.38&
0.758\\

Wan-video V2V &
101.86 &
1607.63 &
\underline{0.9463} &
0.9758 &
0.9821 &
0.7166 &
58.65 &
873.26 &
\underline{0.9360} &
0.9750 &
0.9848 &
0.7899&
15.14 &
0.547 \\

\midrule

\textbf{Ours} &
\underline{76.04} &
\textbf{1290.25} &
\textbf{0.9471} &
\textbf{0.9833} &
\textbf{0.9877} &
0.3817 &
\textbf{35.82} &
\textbf{580.05} &
\textbf{0.9405} &
\textbf{0.9815} &
\textbf{0.9889} &
\textbf{0.3954}&
\textbf{25.47} &
\textbf{0.839} \\

\bottomrule
\end{tabular}%
}
\vspace{-0.4em}
\caption{
\textbf{Quantitative comparison on restoration and harmonization tasks.}
The best and the second best scores are in \textbf{bold} and \underline{underlined}, respectively.
All DS scores are multiplied by 100.
% Bold denotes the best result, and underline denotes the second-best result.
}
\label{tab:exp_main}
\vspace{-0.4em}
\end{table*}

\newcommand{\doublerule}{\midrule\addlinespace[0.05pt]\midrule}
\begin{table*}[t]
\centering

\resizebox{\textwidth}{!}{%
\begin{tabular}{c|ccccc|ccccc|ccccccc}
\toprule
  \multirow{2}{*}[-0.5ex]{\textbf{ID}} &
  \multicolumn{5}{c|}{\textbf{Module}} &
    \multicolumn{5}{c|}{\textbf{Harmonization}} &
    \multicolumn{7}{c}{\textbf{Restoration}} \\
  \cmidrule(lr){2-6} \cmidrule(lr){7-11} \cmidrule(lr){12-18}& 
  \textbf{\begin{tabular}[c]{@{}c@{}}GateNet\end{tabular}}
   &
  \textbf{\begin{tabular}[c]{@{}c@{}}Mask\end{tabular} } &
  \textbf{\begin{tabular}[c]{@{}c@{}}Edge\end{tabular} } &
  \textbf{\begin{tabular}[c]{@{}c@{}}CMask\end{tabular} } &
  \textbf{R-Train} &
  \textbf{FID} $\downarrow$&
  \textbf{FVD} $\downarrow$&
  \textbf{SC} $\uparrow$&
  \textbf{TF} $\uparrow$&
  \textbf{MS} $\uparrow$&
  \textbf{FID} $\downarrow$&
  \textbf{FVD} $\downarrow$&
  \textbf{SC} $\uparrow$&
  \textbf{TF} $\uparrow$&
  \textbf{MS} $\uparrow$&
  \textbf{PSNR} $\uparrow$&
  \textbf{SSIM} $\uparrow$\\
  \midrule
0 &
  &
  &
  &
  &
  &
  84.55 &
  1488.98 &
  0.9424 &
  0.9769 &
  0.9839 &
  44.75 &
  782.38 &
  0.9184 &
  0.9758 &
  0.9849 &
  21.83 &
  0.768 \\

1 &
  \checkmark&
  \checkmark&
  \checkmark&
  \checkmark&
  &
  76.05 &
  1309.77 &
  0.9454 &
  0.9808 &
  0.9858 &
  36.23 &
  594.91 &
  0.9336 &
  0.9814 &
  0.9882 &
  24.93 &
  0.829 \\

2 &
  \checkmark&
  \checkmark&
  \checkmark&
  &
  \checkmark&
  76.61 &
  1295.23 &
  0.9467 &
  0.9832 &
  0.9835 &
  37.28 &
  626.67 &
  0.9328 &
  \textbf{0.9818} &
  0.9859 &
  24.12 &
  0.819 \\

3 &
  \checkmark&
  \checkmark&
  &
  \checkmark&
  \checkmark&
  81.09 &
  1341.10 &
  0.9469 &
  0.9817 &
  0.9864 &
  39.80 &
  640.23 &
  0.9315 &
  0.9787 &
  0.9870 &
  21.94 &
  0.772 \\

4 &
  \checkmark&
  &
  \checkmark&
  \checkmark&
  \checkmark&
  80.43 &
  1324.53 &
  0.9439 &
  0.9798 &
  0.9851 &
  37.24 &
  584.53 &
  0.9351 &
  0.9803 &
  0.9875 &
  22.97 &
  0.794 \\

5 &
  &
  \checkmark&
  \checkmark&
  \checkmark&
  \checkmark&
  79.42 &
  1334.83 &
  0.9460 &
  0.9814 &
  0.9873 &
  37.29 &
  598.31 &
  0.9335 &
  0.9798 &
  0.9871 &
  24.23 &
  0.817 \\

6 &
  \checkmark&
  \checkmark&
  \checkmark&
  \checkmark&
  \checkmark&
  \textbf{76.04} &
  \textbf{1290.25} &
  \textbf{0.9471} &
  \textbf{0.9833} &
  \textbf{0.9877} &
  \textbf{35.82} &
  \textbf{580.05} &
  \textbf{0.9405} &
  0.9815 &
  \textbf{0.9889} &
  \textbf{25.47} &
  \textbf{0.839} \\

  \bottomrule

\end{tabular}%

}
\vspace{-0.4em}
\caption{
\textbf{Ablation study of guidance modules and reward-guided post-training.}
We ablate GateNet, foreground mask guidance (\textbf{Mask}), edge-map constraint (\textbf{Edge}), category-level mask guidance (\textbf{CMask}), and reward-guided post-training (\textbf{R-Train}).
}
\vspace{-1.2em}
\label{tab:abl_app}
\end{table*}

\begin{table}[t]
\centering
\small
\setlength{\tabcolsep}{6pt}
\begin{tabular*}{\linewidth}{@{\extracolsep{\fill}} l l c @{}}
\toprule
\textbf{Study} & \textbf{Method} & \textbf{Result} \\
\midrule

\multirow{4}{*}{Preference Study}
& Harmonizer & 88.3\% \\
& Fixer & 90.2\% \\
& Cosmos-Transfer2.5 & 92.4\% \\
& Wan-video V2V & 95.7\% \\

\bottomrule
\end{tabular*}
\vspace{-0.4em}
\caption{
\textbf{User study results.} A rate above 50\% indicates a preference for RealityBridge over the corresponding baseline.
}
\vspace{-1.2em}
\label{tab:user_study}
\end{table}

\subsection{Experimental Setting}

\textit{\textbf{RealityBridge}} uses Wan2.2-Fun-VACE-A14B~\cite{vace} as its video backbone. We provide detailed model, training, and inference settings in Appendix~\ref{app:experimental_details}.

\paragraph{Evaluation Protocol.}
We evaluate the {RealityBridge} model on two tasks: restoration of degraded 3DGS-rendered videos and harmonization after 3D asset insertion. Our evaluation includes 1K internal scenes disjoint from the training set, 60 Waymo scenes~\cite{waymo}, and 10 NuPlan scenes~\cite{nuplan}. For restoration, degraded inputs follow the 3DGS Artifact-to-Real construction described in the data-curation section, with the corresponding real videos used as aligned references. For insertion, we use 3D assets in the internal set and Hunyuan3D-reconstructed vehicles~\cite{hunyuan-3d} in public scenes. 
All test videos are rendered at $720\times1280$, 30 FPS, and 10 seconds.
%All test videos are rendered at $720\times1280$ resolution, 30 FPS, and 10 seconds in duration. 
% RealityBridge is built upon Wan2.2-Fun-VACE-A14B~\cite{vace}; detailed settings are provided in the appendix.

\paragraph{Baselines.}
We compare RealityBridge with four baseline groups: the general image editing models SDEdit~\cite{meng2021sdedit} and InstructPix2Pix~\cite{brooks2023instructpix2pix}, the video harmonization method Harmonizer~\cite{diff-harmonizer} for asset insertion, general video translation models Cosmos-Transfer2.5~\cite{cosmos} and Wan2.2-Fun-VACE-A14B~\cite{vace} (referred to as Wan-video V2V), and the 3DGS artifact removal method Fixer. Detailed settings are provided in Appendix~\ref{app:experimental_details}.

\paragraph{Evaluation Metrics.}
We evaluate visual realism using {FID}~\cite{fid} and {FVD}~\cite{fvd}, which measure the distribution distance between generated and real driving videos. Temporal stability is assessed using {VBench}~\cite{vbench} metrics, including Subject Consistency (SC), Temporal Flickering (TF), and Motion Smoothness (MS). For paired settings with references, we also report {PSNR}~\cite{psnr} and {SSIM}~\cite{ssim} for reconstruction fidelity; for unpaired asset insertion, where no ground-truth video (GT) exists for inserted objects, full-reference metrics are omitted as primary measures. 
To assess structure preservation, we report DINOv2 Structure Distance (DS)~\cite{Tumanyan_2022_CVPR}, comparing outputs with GT for restoration and with input composites for harmonization, where harmonized GT is unavailable.
% We further measure scene structure preservation using DINOv2~\cite{dinov2} feature similarity (DS), computed between generated video and input video for harmonization, and between generated video and groundtruth video for restoration.

\subsection{Qualitative Evaluation}

We present qualitative comparisons on restoration and inserted-asset harmonization in Fig.~\ref{fig:qualitative_eval}. For restoration, Cosmos-Transfer2.5, Fixer and Harmonizer fail to sufficiently restore degraded vehicles and remove local artifacts, while Wan-video V2V tends to over-modify the input appearance and geometry. In contrast, RealityBridge produces clean vehicle appearances with fewer residual artifacts while preserving the original scene layout and road structure.
For inserted-asset harmonization, inserted vehicles should be visually compatible with the environment while remaining faithful to the input geometry and trajectory. Cosmos-Transfer2.5 produces insufficient cast shadows, Fixer yields weak shadow effects, Harmonizer produces faint, overlay-like shadows that are poorly integrated with the scene, and Wan-video V2V often over-modifies the appearance and geometry of the inserted assets. RealityBridge achieves more realistic vehicle-environment harmonization with coherent lighting, shadows, and boundary blending while preserving the input structure and motion. Moreover, the consistent performance on both internal and public datasets further validates the applicability of RealityBridge across different scene distributions, with more visual results and multi-view results from different camera types provided in Appendix~\ref{app:qualitative}. Additional video results are provided on the project page.

\subsection{Quantitative Evaluation}
Tab.~\ref{tab:exp_main} reports the quantitative results. \textit{On the harmonization task}, RealityBridge outperforms baselines across most video-level and temporal consistency metrics. These results demonstrate that RealityBridge improves foreground-background integration while maintaining temporally coherent appearance across frames. Although several baselines obtain competitive frame-level scores, their weaker FVD and temporal consistency results indicate limited video-level realism and inter-frame coherence. 
% For DS, RealityBridge achieves competitive results, while higher scores from some baselines are attributed to weaker editing that preserves more input appearance.
Regarding DS, our slightly higher score reflects stronger relighting and shadow synthesis, which the input-referenced metric may penalize, while conservative methods benefit from remaining closer to input.

\textit{On the 3DGS restoration task}, RealityBridge shows clear advantages across perceptual, temporal, and reference-based metrics compared to existing methods, with notable gains in FID, FVD and DS. This confirms its ability to correct rendering artifacts, restore degraded foreground regions and boundaries, and preserve temporal stability.

We further conduct a user study with 50 participants to evaluate overall perceptual quality, as reported in Tab.~\ref{tab:user_study}. Participants perform pairwise comparisons between RealityBridge
and four representative baselines and select the preferred result. RealityBridge consistently receives higher preference rates across all pairwise comparisons, confirming its overall perceptual advantage over existing methods. Further details are provided in Appendix~\ref{app:experimental_details}.

\subsection{Ablation Study}

We report the ablation results in Tab.~\ref{tab:abl_app}. Removing any individual component degrades performance on the harmonization and restoration tasks, with different variants exhibiting distinct drops across evaluation metrics. These results indicate that the proposed control mechanisms and post-training strategy provide complementary benefits. The full model achieves the best or second-best performance on all metrics, covering video-level realism, temporal consistency, and reference-based restoration quality. Additional data ablation results are provided in Appendix~\ref{app:qualitative}.

\section{Conclusion}

We presented \textbf{RealityBridge}, a video-level post-rendering refinement framework that converts edited 3DGS driving simulations into photorealistic videos. 
RealityBridge integrates a video-based refinement model with adaptive multimodal control, a targeted data curation pipeline, and reward-guided post-training, which jointly address rendering artifacts, degraded foreground assets, illumination mismatches, missing shadows, and temporal flickering.
Extensive experiments demonstrate consistent improvements over existing methods in 3DGS restoration, inserted-asset harmonization, and temporal consistency. 
We believe RealityBridge offers a practical solution for photorealistic simulation in autonomous driving and enables the generation of high-quality data that can support downstream driving tasks.

{
\small
\bibliographystyle{ieeenat_fullname}
\bibliography{main,supp-extra}
}

\clearpage
\setcounter{page}{1}
\appendix
\section*{Appendix Overview}

\begin{itemize}
    \item \textbf{Section~\ref{app:data_curation}}: Data Curation Details.
    \item \textbf{Section~\ref{app:training_strategy}}: Training Strategy Details.
    \item \textbf{Section~\ref{app:experimental_details}}: Experimental Details.
    \item \textbf{Section~\ref{app:qualitative}}: Qualitative Examples.
\end{itemize}

\section{Data Curation Details}
\label{app:data_curation}

This section provides additional details on our targeted data curation pipeline for 3DGS-to-Real supervision. We curate paired data for four major failure modes: rendering artifacts, illumination mismatch, human-region degradation, and small-object detail loss, with aligned multimodal conditions for controllable restoration. 

\subsection{Paired Supervision Construction}

\paragraph{3DGS Artifact-to-Real Pairs.}
To provide supervision for repairing degradation, we construct 3DGS Artifact-to-Real pairs from internal multi-view driving logs covering diverse traffic environments, including highways, urban roads, garages, commercial streets, residential neighborhoods, and rural field roads. Adapting the data strategy of Difix3D+~\cite{difix3d}, we generate degraded renderings using four reconstruction protocols: \textbf{\textit{sparse reconstruction}}, \textbf{\textit{cycle reconstruction}}, \textbf{\textit{cross-camera reference}}, and \textbf{\textit{underfitting}}. \textit{(i) Sparse reconstruction} optimizes a 3DGS scene from temporally subsampled observations and renders held-out frames along the corresponding trajectory as degraded inputs. \textit{(ii) Cycle reconstruction} renders a laterally perturbed trajectory from an initial 3DGS scene, reconstructs a second 3DGS scene from these pseudo observations, and renders back the original camera trajectory as degraded inputs. \textit{(iii) Cross-camera reference} reconstructs the scene from one camera view and renders another camera view, introducing view-dependent artifacts and geometric inconsistency. \textit{(iv) Region-aware underfitting} stops reconstruction early to synthesize blur, floaters, and incomplete geometry. We apply different underfitting levels to foreground assets and background scenes, mimicking the stronger degradation often seen on inserted or dynamic objects. Original real videos serve as supervision targets, enabling the model to learn artifact removal while preserving original scene layout and camera motion.

\paragraph{Illumination Variation.}
To supervise foreground harmonization, we construct illumination-variation clips from real videos using two strategies: \textit{\textbf{RGB-level perturbation}} and \textit{\textbf{normal-guided relighting}}. 
For \textit{RGB-level perturbation}, we modify foreground regions by randomly adjusting brightness, contrast, gamma, and per-channel color gains within moderate ranges to simulate appearance mismatch introduced by asset insertion.
For \textit{normal-guided relighting}, we estimate temporally consistent surface normals with NormalCrafter~\cite{normal-crafter} and sample random light conditions inspired by Lumen~\cite{lumen}. Specifically, for each foreground pixel $x$ at time $t$, we use its estimated normal $n_t(x)$ to compute a normal-dependent illumination map from a set of time-varying light sources $\mathcal{P}_t$:
\begin{equation}
    L_t(x) =
    \sum_{p \in \mathcal{P}_t}
    I_p^{(t)}
    \max\left(\langle n_t(x), d_p^{(t)} \rangle, 0\right)
\end{equation}
where $I_p^{(t)}$ and $d_p^{(t)}$ denote the RGB intensity and direction of light source $p$, respectively. The inner product $\langle n_t(x), d_p^{(t)} \rangle$ corresponds to the cosine of the angle between the surface normal and the incident light direction, so surfaces facing the sampled light receive stronger illumination. 
The resulting illumination map is then applied to the original appearance:
\begin{equation}
\widetilde{I}_t(x)
=
\operatorname{clip}_{[0,255]}
\left(
\frac{I_t(x)}{255}
\odot L_t(x)
\right),
\end{equation}
where $I_t(x)\in[0,255]^3$ is the original RGB value at pixel $x$, $\widetilde{I}_t(x)$ is its relit counterpart, and $\odot$ denotes channel-wise multiplication. Clipping is performed independently for each RGB channel.
Both strategies are applied only within the foreground region, while the original video is retained as the paired target. This subset provides supervision for correcting foreground illumination mismatch and improving foreground-background boundary blending.

\paragraph{Human-Motion Prior.}
To improve robustness on pedestrian regions, we curate human-motion prior clips from an internal pedestrian-dominant dataset. These videos cover diverse human motions, large pose changes, occlusions, and multi-person interactions, providing pedestrian-centric observations often missing in ordinary driving data. These videos are captured by in-vehicle cameras when the ego vehicle is static to reduce ego-motion blur and make human-region degradation more controllable.
We synthesize degraded counterparts using the degradation pipeline from Real-ESRGAN~\cite{realesrgan} along with illumination perturbation to enhance robustness in human-region restoration. Specifically, we apply a two-stage degradation pipeline with random blur, resizing, Gaussian or Poisson noise, JPEG compression, video compression, and optional resampling/filtering operations. Degraded clips are resized back to the original resolution and paired with the original clips as clean supervision.

\paragraph{Small-Object Asset Videos.}
To improve preservation of small critical objects, we curate object-centric clips from our internal 3D asset pipeline. These clips contain 360-degree captures of traffic signs, cones, barriers, fallen objects, and other small obstacles that are common but often low-salience in driving videos and easily lost during 3DGS reconstruction or edited rendering. When degraded supervision is needed, we process these clips with the same degradation pipeline. The clean asset videos are used as supervision targets for small-object restoration and preservation.

\subsection{Multimodal Condition Generation}

For each paired sample, we construct multimodal conditions to enable fine-grained control. To preserve global scene context, we use the processed video as the visual condition, which provides layout, camera motion, and appearance cues. 
To localize regions requiring stronger restoration or harmonization, we use Segment Anything Model 3 (SAM3)~\cite{sam3} to extract two types of masks. First, foreground masks identify the assets to be modified, primarily including vehicles, pedestrians, and small objects that are inserted, edited, or degraded in the input video. Second, category-level masks encode broader scene semantics by distinguishing foreground assets and additional categories such as traffic signs, barriers, and background regions, thereby providing semantic priors for generation.

To preserve safety-critical structures, we extract edge maps using the Canny~\cite{canny} operator from the OpenCV library~\cite{opencv}. These edge maps provide boundary-level guidance for object contours, lane markings, traffic signs, and other fine structures. For language-level scene guidance, we use Qwen3-VL-8B~\cite{qwen3vl} to generate captions from uniformly sampled video frames. The captioning prompt encourages the model to describe the scene type, foreground objects, degradation pattern, and expected restoration behavior, providing complementary semantic guidance for controllable video restoration and harmonization.

\section{Training Strategy Details}
\label{app:training_strategy}

\begin{figure}[t]
    \centering
    \includegraphics[width=1.0\linewidth]{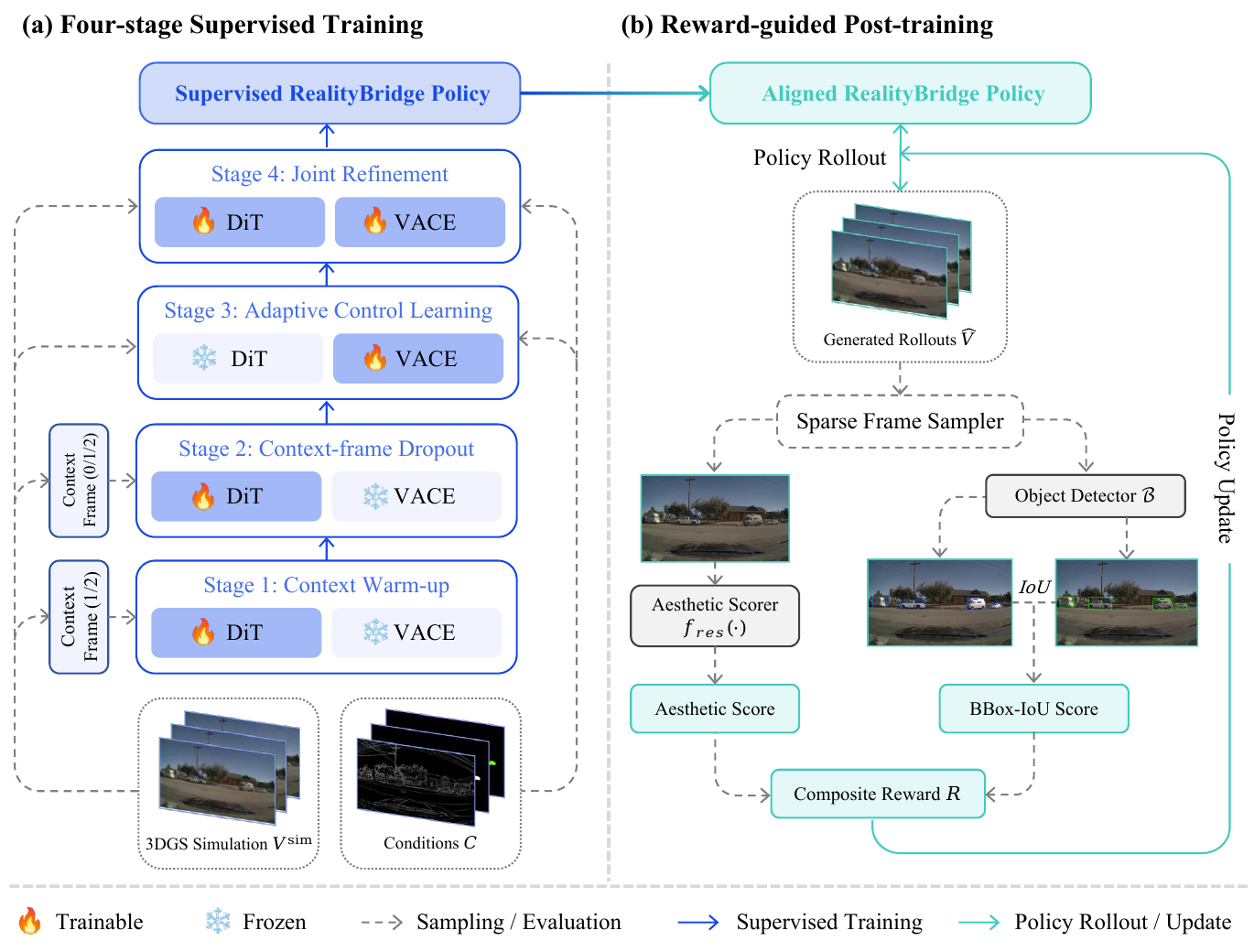}
    \vspace{-1.2em}
    \caption{
    \textbf{Training pipeline.}
    The training procedure of the RealityBridge framework consists of (a) four-stage supervised training and (b) reward-guided post-training.
    }
    \vspace{-1.2em}
    \label{fig:training_strategy}
\end{figure}

This section provides additional details of the training strategy used in RealityBridge, including the {{four-stage supervised training}} curriculum and the {{reward-guided post-training}} stage. As shown in Fig.~\ref{fig:training_strategy}, the supervised curriculum progressively develops the model's capabilities for temporal continuation and multimodal control, while reward-guided post-training further promotes photorealistic generation under simulator-defined constraints.

\subsection{Four-stage Supervised Training}

The supervised training stage supports chunk-wise autoregressive processing. Instead of training all modules jointly from the beginning, we progressively introduce context-conditioned chunk prediction, context-frame dropout, multimodal condition learning, and joint refinement.

\begin{figure*}[!thb]
    \centering
    \includegraphics[width=0.92\textwidth]{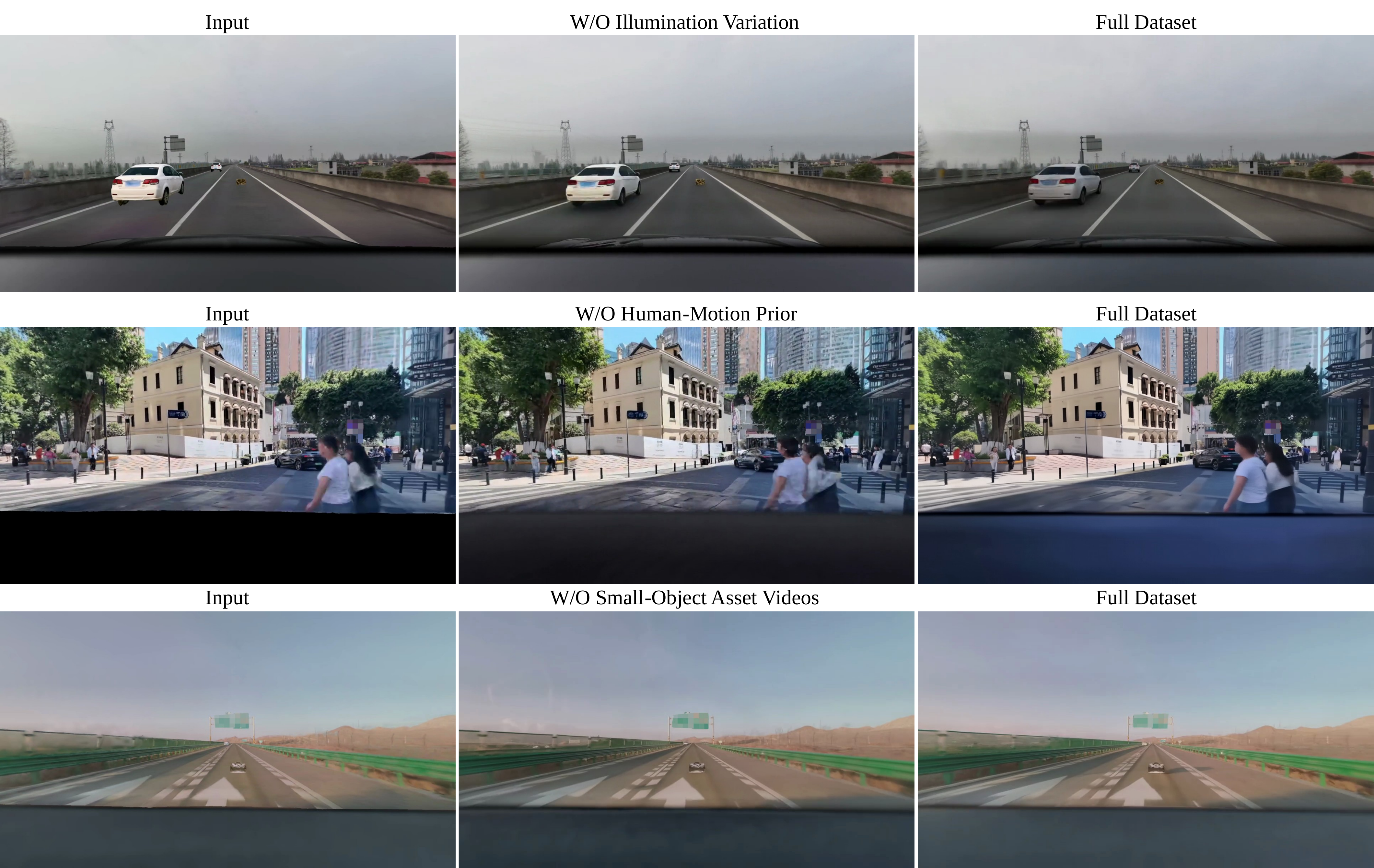}
    \vspace{-0.2em}
    \caption{
    \textbf{Qualitative ablation of curated training data.}
    Each row presents the input alongside results from an ablated variant and the model trained on the full dataset. The top, middle, and bottom rows examine the effects of illumination-variation data, human-motion prior data, and small-object asset videos, respectively.
    }
    \vspace{-1.2em}
    \label{fig:supp_data_ablation}
\end{figure*}

\begin{itemize}
\item[$\bullet$]\textbf{Stage 1: Context warm-up.}
During training, the first one or two noisy latent frames are replaced with the corresponding clean latent frames encoded from the target real video, and the model is trained to predict the remaining target-video latents.
This stage builds basic temporal continuation from limited observations.

\item[$\bullet$]\textbf{Stage 2: Context-frame dropout.}
We then continue training the DiT backbone with context-frame dropout, where the number of context frames is sampled from $\{0,1,2\}$. The zero-frame case simulates first-chunk initialization without historical context, while the one- and two-frame cases improve context-conditioned continuation for subsequent chunks.

\item[$\bullet$]\textbf{Stage 3: Adaptive control learning.}
Next, we freeze the DiT backbone and train the ControlNet branch together with GateNet. This stage learns to convert multimodal conditions into effective control features and adaptively allocate them across different DiT blocks.

\item[$\bullet$]\textbf{Stage 4: Joint refinement.}
Finally, we jointly fine-tune the DiT backbone, ControlNet and GateNet. This stage combines context-conditioned chunk continuation with multimodal conditional control for chunk-wise autoregressive restoration and harmonization.
\end{itemize}

\begin{figure*}[!thb]
    \centering
    \includegraphics[width=0.88\textwidth]{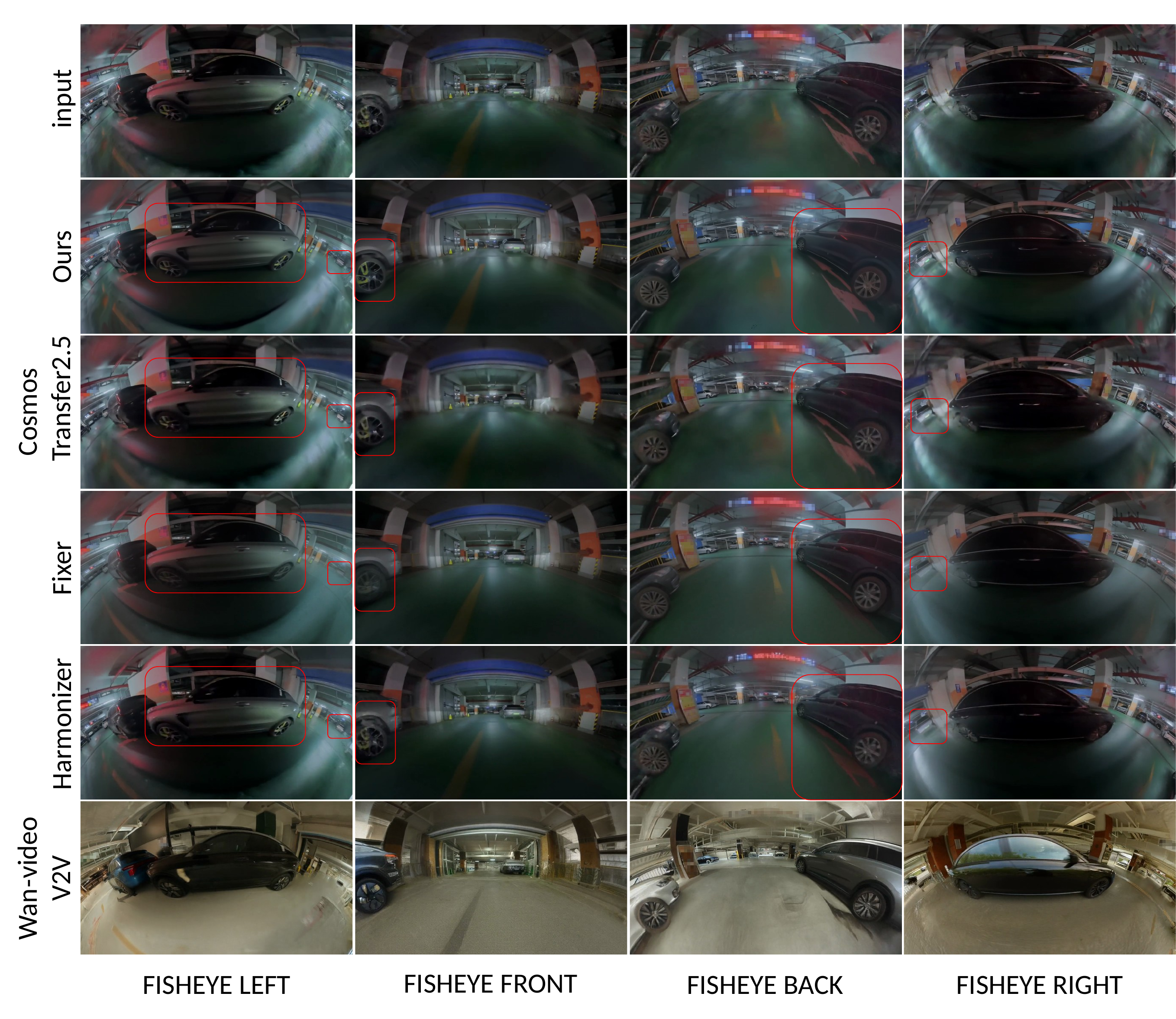}
    \vspace{-0.6em}
    \caption{
    \textbf{Cross-camera robustness comparison.}
    Our method and baselines are evaluated independently on fisheye left, front, back, and right camera videos from the same scene. Boxes highlight regions where restoration and harmonization differ most clearly.
    }
    \vspace{-1em}
    \label{fig:supp_multiview}
\end{figure*}

\subsection{Reward-guided Post-training}

After supervised training, we apply reward-guided post-training as a final alignment step. This stage further improves perceptual realism while suppressing foreground drift and hallucinated details under simulator constraints. 

\paragraph{Aesthetic Reward.}
To encourage realistic appearance, coherent motion, and text-consistent
restoration, we follow the preference-training paradigm of
Flow-DPO~\cite{flow-dpo} and train an aesthetic reward model based on
Qwen3-VL-2B~\cite{qwen3vl}. Given the sampled frame set
$\{\hat{I}_{t}\}_{t\in\mathcal{S}}$, the reward model jointly predicts
three scalar scores for \textit{visual quality}, \textit{motion quality},
and \textit{text alignment}. The aesthetic reward is computed as:
\begin{equation}
\begin{aligned}
f_{\mathrm{aes}}\left(\{\hat{I}_{t}\}_{t\in\mathcal{S}}\right)
&=
\alpha_{\mathrm{v}}
s_{\mathrm{v}}\left(\{\hat{I}_{t}\}_{t\in\mathcal{S}}\right) \\
&\quad+
\alpha_{\mathrm{m}}
s_{\mathrm{m}}\left(\{\hat{I}_{t}\}_{t\in\mathcal{S}}\right)
+
\alpha_{\mathrm{t}}
s_{\mathrm{t}}\left(\{\hat{I}_{t}\}_{t\in\mathcal{S}}\right),
\end{aligned}
\end{equation}
where $s_{\mathrm{v}}(\cdot)$, $s_{\mathrm{m}}(\cdot)$, and $s_{\mathrm{t}}(\cdot)$ are the
visual-quality, motion-quality, and text-alignment scores jointly
predicted by the reward model, respectively, and
$\alpha_{\mathrm{v}}$, $\alpha_{\mathrm{m}}$, and
$\alpha_{\mathrm{t}}$ are their corresponding weighting coefficients.

\paragraph{Foreground Structural Reward.}
To penalize foreground spatial drift during realism enhancement, we use YOLOv11~\cite{yolov11} as the object detector $\mathcal{B}(\cdot)$ to obtain foreground bounding boxes from the generated frame $\hat{I}_{t}$ and the rendered input frame $I^{r}_{t}$. For a detected box $\hat{b}_{t}$ in $\hat{I}_{t}$ and its matched box $b^{r}_{t}$ in $I^{r}_{t}$, the bounding-box IoU is computed as:
\begin{equation}
    \operatorname{IoU}(\mathcal{B}(\hat{I}_{t}), \mathcal{B}(I^{r}_{t})) = \operatorname{IoU}(\hat{b}_{t}, b^{r}_{t})
    =
    \frac{
    \left|\hat{b}_{t}\cap b^{r}_{t}\right|
    }{
    \left|\hat{b}_{t}\cup b^{r}_{t}\right|
    }.
\end{equation}
When multiple foreground objects are detected, we match boxes with the same object category and average their IoU scores over valid matches. This term encourages the generated video to preserve foreground object locations and suppress spatial drift during realism enhancement.

\section{Experimental Details}
\label{app:experimental_details}

\paragraph{Implementation Details.}
We train at $720\times1280$ resolution on 8 NVIDIA H200 GPUs with a learning rate of $1\times10^{-6}$. The first two training stages are run for 20K iterations each, and the last two stages for 10K iterations each. We set $\lambda_{\mathrm{reg}}=0.1$ for regional reweighting and $\lambda_{\mathrm{aes}}=\lambda_{\mathrm{box}}=1$ for reward-guided post-training. During inference, videos exceeding a single model window are processed autoregressively in consecutive chunks.

\paragraph{Baseline Implementation.}
All baselines are evaluated on the same input videos and at the same output resolution. We use official implementations and released checkpoints whenever available. SDEdit~\cite{meng2021sdedit} uses Stable Diffusion~3 as the denoising backbone and performs image-to-image translation by perturbing each input frame with noise and subsequently denoising it. InstructPix2Pix~\cite{brooks2023instructpix2pix} performs instruction-guided image editing conditioned on both the input frame and a task-specific text instruction. Both image-based baselines are applied frame by frame. Cosmos-Transfer2.5~\cite{cosmos} is evaluated using visual and edge conditions. Wan-video V2V uses Wan2.2-Fun-VACE-A14B~\cite{vace}, with the input video provided as the video-to-video condition. Harmonizer~\cite{diff-harmonizer} is evaluated following its official setup. For 3DGS artifact removal, we use Fixer~\cite{difix3d} and follow its official setup, treating it as a single-step image diffusion enhancer for rendered novel views and applying it frame by frame to the 3DGS-rendered videos.

\paragraph{User Study.}
We conduct a user study with 50 internal participants from multiple groups and diverse professional backgrounds, so that the evaluation reflects a broader range of visual judgment and domain familiarity. Each participant evaluates 100 videos on average. For each input scene, participants compare RealityBridge with one baseline at a time under the same input condition and select the result with better overall Sim-to-Real quality. The judgment considers visual realism, clean artifact-free appearance, foreground-background harmonization, preservation of edited structures, and temporal stability. Method order is randomized to reduce presentation bias. We directly report the preference rates without confidence intervals.

% \begin{figure*}[t!]
%     \centering
%     \includegraphics[width=0.94\textwidth]{figures/appendix_grid1.pdf}
%     \caption{
%         \textbf{Additional qualitative comparisons on harmonization and restoration.}
%         We compare RealityBridge with Cosmos-Transfer2.5, Fixer, Harmonizer, and Wan-video V2V across sampled video frames.
%     }
%     \label{fig:supp_video_results_1}
% \end{figure*}

\section{Qualitative Examples}
\label{app:qualitative}

\subsection{Qualitative Data Ablation}
We visualize the contribution of targeted data curation in Fig.~\ref{fig:supp_data_ablation}. The ablation focuses on curated subsets with clear localized effects: illumination-variation clips, human-motion prior clips, and small-object asset videos. Removing illumination-variation data weakens the model's ability to correct contact shadows and vehicle-surface lighting, while removing human-motion prior data leads to poorer pedestrian-region restoration, especially under large pose changes or partial occlusions. For small objects, removing the asset videos causes harmonization to largely fail: the result remains close to the input rendering and lacks plausible light-shadow integration with the scene. The full model produces more coherent small-object appearance and local lighting. These qualitative results show that each curated subset addresses a distinct failure mode.

% \begin{figure*}[t!]
%     \centering
%     \includegraphics[width=0.93\textwidth]{figures/appendix_grid2.pdf}
%     \caption{
%         \textbf{Additional qualitative comparisons on harmonization and restoration.}
%         We compare RealityBridge with Cosmos-Transfer2.5, Fixer, Harmonizer, and Wan-video V2V across sampled video frames.
%     }
%     \label{fig:supp_video_results_2}
% \end{figure*}

\subsection{Cross-camera Robustness}
Driving simulators often render videos from cameras with different viewing directions and lens distortions. To evaluate cross-camera robustness, we compare results on four fisheye camera videos from the same scene, covering left, front, back, and right views. As shown in Fig.~\ref{fig:supp_multiview}, our method maintains stable restoration and harmonization quality across these camera settings, while baselines are more sensitive to view direction, fisheye distortion, and local 3DGS artifacts.

\subsection{Additional Visual Results}
\begin{figure*}[p]
    \centering
    \includegraphics[width=\linewidth]{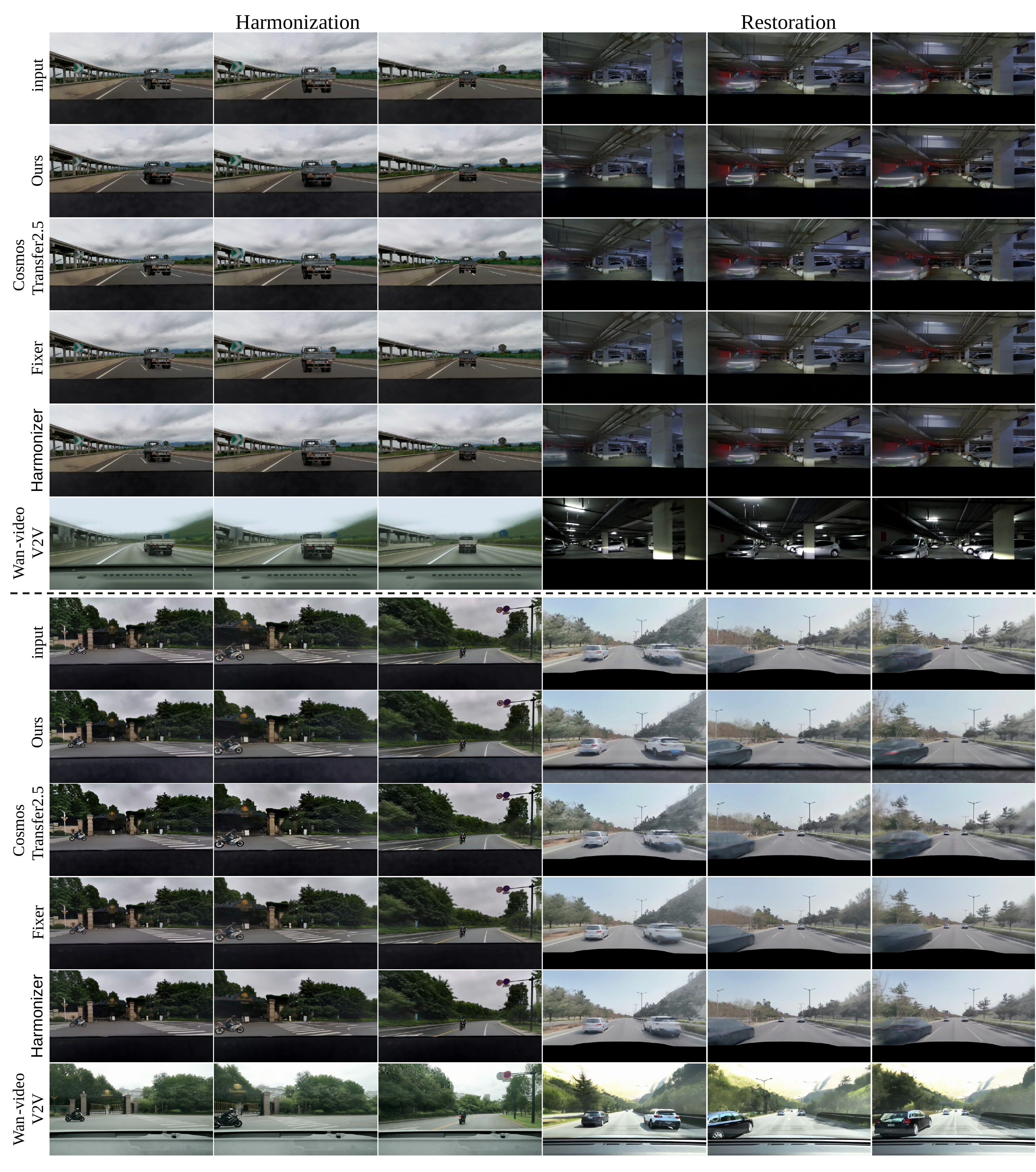}
    \caption{
        \textbf{Additional qualitative comparisons on harmonization and restoration.}
        We compare RealityBridge with Cosmos-Transfer2.5, Fixer, Harmonizer, and Wan-video V2V across sampled video frames.
    }
    \label{fig:supp_video_results_1}
\end{figure*}

\begin{figure*}[p]
    \centering
    \includegraphics[width=\linewidth]{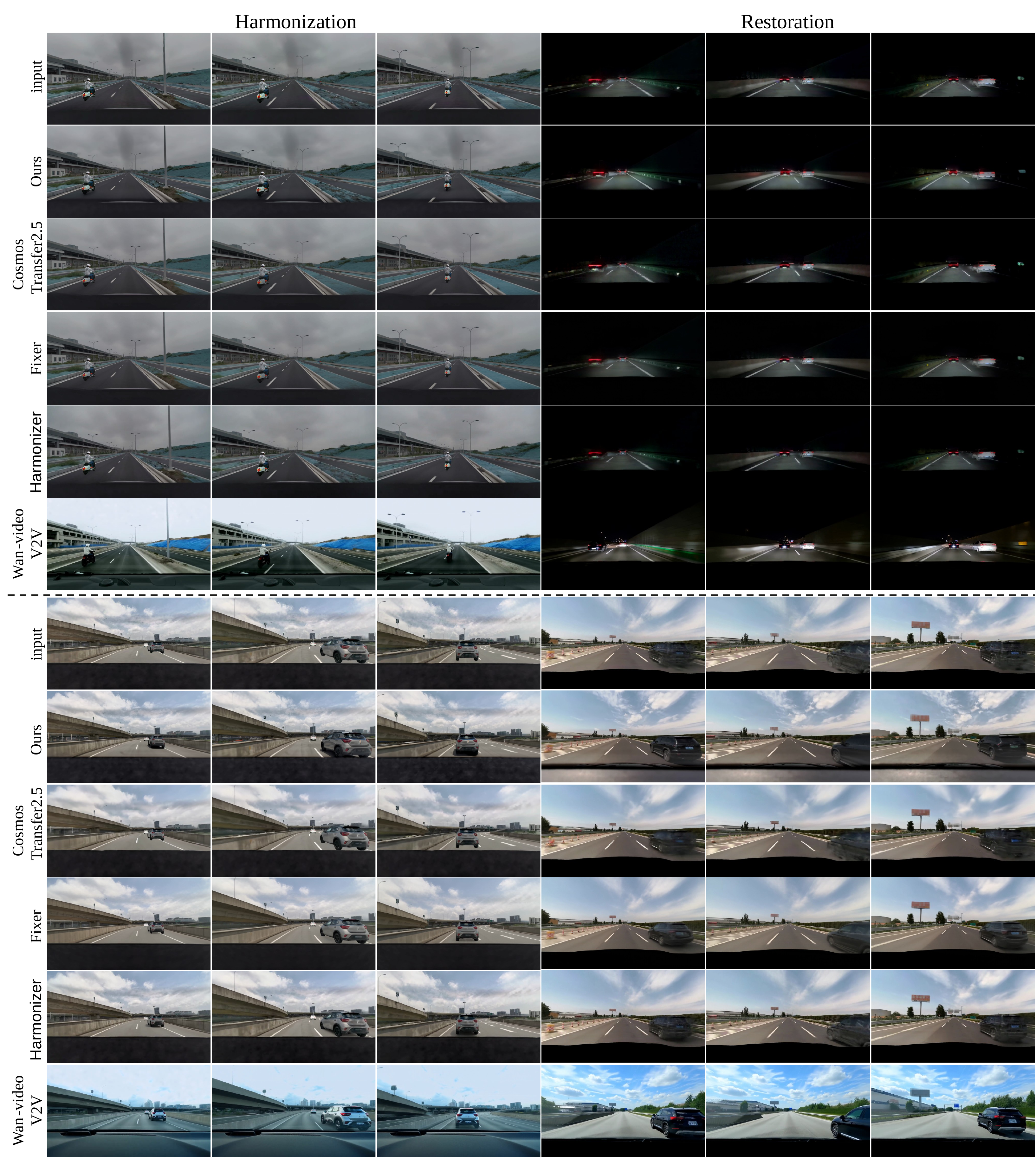}
    \caption{
        \textbf{Additional qualitative comparisons on harmonization and restoration.}
        We compare RealityBridge with Cosmos-Transfer2.5, Fixer, Harmonizer, and Wan-video V2V across sampled video frames.
    }
    \label{fig:supp_video_results_2}
\end{figure*}

We provide additional qualitative examples in Fig.~\ref{fig:supp_video_results_1} and Fig.~\ref{fig:supp_video_results_2} to complement the main comparison.

\end{document}